\documentclass[11pt]{article}
\usepackage{eacl2017}
\usepackage{times}
\usepackage{url}
\usepackage{latexsym}
\usepackage{float,tabularx}

\usepackage[utf8]{inputenc}

\usepackage{graphicx}

\eaclfinalcopy 

\title{Dialectometric analysis of language variation in Twitter}

\author{Gonzalo Donoso \\
  IFISC (UIB-CSIC) \\
  Palma de Mallorca, Spain \\
  {\tt gdonoso94@hotmail.com} \\\And
  David S\'anchez \\
  IFISC (UIB-CSIC) \\
  Palma de Mallorca, Spain \\
  {\tt david.sanchez@uib.es} \\}

\date{}

\begin{document}
\maketitle
\begin{abstract}
In the last few years, microblogging platforms such as Twitter
have given rise to a deluge of textual data that can be used for the analysis of informal communication between millions of individuals.
In this work, we propose an information-theoretic approach to geographic language variation using a corpus based on Twitter.
We test our models with tens of concepts and their associated keywords detected in Spanish tweets geolocated in Spain.
We employ dialectometric measures (cosine similarity and Jensen-Shannon divergence)
to quantify the linguistic distance on the lexical level between cells created in a uniform grid over the map.
This can be done for a single concept or in the general case taking into account an average of the considered
variants. The latter permits an analysis of the dialects that naturally emerge from the data.
Interestingly, our results reveal the existence of two dialect macrovarieties. 
The first group includes a region-specific speech spoken in small towns
and rural areas whereas the second cluster encompasses cities that tend to use a more uniform variety.
Since the results obtained with the two different metrics qualitatively agree,
our work suggests that social media corpora can be efficiently used
for dialectometric analyses.
\end{abstract}

\section{Introduction}

Dialects are language varieties defined across space. These varieties can differ in distinct linguistic levels
(phonetic, morphosyntactic, lexical), which determine a particular regional speech~\cite{chambers98-1}.
The extension and boundaries (always diffuse) of a dialect area are obtained from the variation of one or many
features such as, e.g., the different word alternations for a given concept. Typically, the dialect forms plotted
on a map appear as a geographical continuum that gradually connects places with slightly different diatopic
characteristics. A dialectometric analysis aims at a computational approach to dialect distribution, providing
quantitative linguistic distances between locations~\cite{seg71,goe06,wie15}.

Dialectometric data is based upon a corpus that contains the linguistic information needed for the statistical
analysis. The traditional approach is to generate these data from surveys and questionnaires that
address variable types used by a few informants. Upon appropriate weighting, the distance metric can thus be mapped
on an atlas. In the last few years, however, the impressive upswing of microblogging platforms has led to a scenario in which human communication features can be studied without the effort that traditional studies usually require. Platforms such as Twitter, Flickr, Instagram or Facebook bring us the possibility of investigating massive amounts of data in an automatic fashion.
Furthermore, microblogging services provide us with real-time communication among users that, importantly,
tend to employ an oral speech.
Another difference with traditional approaches is that while the latter focus on male, rural informants, users of social platforms
are likely to be young, urban people \cite{twitterusers}, which opens the route to novel investigations on today's usage of language.
Thanks to advances in geolocation, it is now possible to directly examine
the diatopic properties of specific regions. Examples of computational linguistic works that investigate regional
variation with Twitter or Facebook corpora thus far comprise English \cite{eisenstein14-1,doyle14-1,kulkarni16-1,huang16-1,blo16}, Spanish \cite{BD,goncalves16-1,malmasi-EtAl:2016:VarDial3}, German \cite{sch14}, Arabic \cite {lin14} and Dutch \cite{tul16}.
It is noticeable that many of these works combine big data techniques with probabilistic tools or machine learning strategies
to unveil linguistic phenomena that are absent or hard to obtain from conventional methods (interviews, hand-crafted corpora, etc.).

The subject of this paper is the language variation in a microblogging platform using dialectrometric measures.
In contrast to previous works, here we precisely determine the linguistic distance between different places
by means of two metrics. Our analysis shows that the results obtained with both metrics are compatible,
which encourages future developments in the field. We illustrate our main findings with a careful analysis
of the dialect division of Spanish. For definiteness, we restrict ourselves to Spain but the method can be straightforwardly
applied to larger areas. We find that, due to language diversity, cities and main towns have similar linguistic distances
unlike rural areas, which differ in their homogeneous forms. 
but obtained with a completely different method 

\section{Methods}
Our corpus consists of approximately 11 million geotagged tweets produced in Europe in Spanish language between October 2014
and June 2016. (Although we will focus on Spain, we will not consider in this work the speech of the Canary Islands due to difficulties with the data extraction). The classification of tweets is accomplished by applying the Compact Language Detector (CLD) \cite{CLD} to our dataset. CLD exhibits accurate benchmarks and is thus good for our purposes, although a different detector might be used \cite{langid}.
We have empirically checked that when CLD determines the language with a probability of at least 60\% the results are extremely reliable. Therefore, we only take into account those tweets for which the probability of being written in Spanish is greater than $0.6$. 
Further, we remove unwanted characters, such as hashtags or at-mentions, using \texttt{Twokenize} \cite{twokenizer},
a tokenizer designed for Twitter text in English, adapted to our goals. 

\begin{figure}[t]
\centering
\includegraphics[width=.48\textwidth,clip]{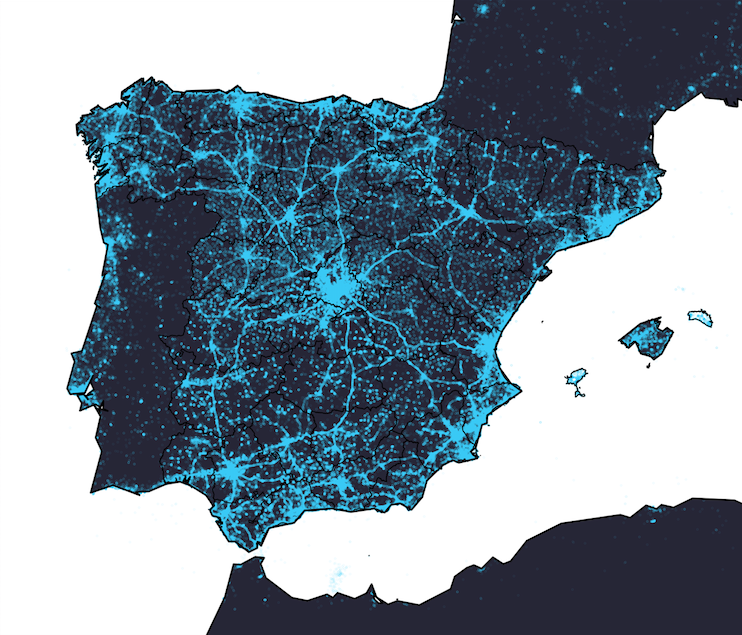}
\caption{Heatmap of Spanish tweets geolocated in Europe. There exist 11208831 tweets arising from a language detection and tokenization procedure. We have zoomed in those arising in Spain, Portugal and the south of France.}
\label{all_tw_es}
\end{figure}

We present the spatial coordinates of all tweets in figure~\ref{all_tw_es}
(only the south-western part of Europe is shown for clarity).
As expected, most of the tweets are localized in Spain, mainly around major cities and along main roads.

Next, we select a word list from \emph{Varilex} \cite{Varilex}, a lexical database that contains Spanish variation across the world. We consider 89 concepts expressed in different forms. Our selection eliminates possible semantic ambiguities. The complete list of keywords is included in the supplementary material below. For each concept, we determine the coordinates of the tweets in which the different keywords appear. From our corpus, we find that 219362 tweets include at least one form corresponding to any of the selected concepts. 

\begin{figure*}[t]
\centering
{\includegraphics[width=.45\textwidth]{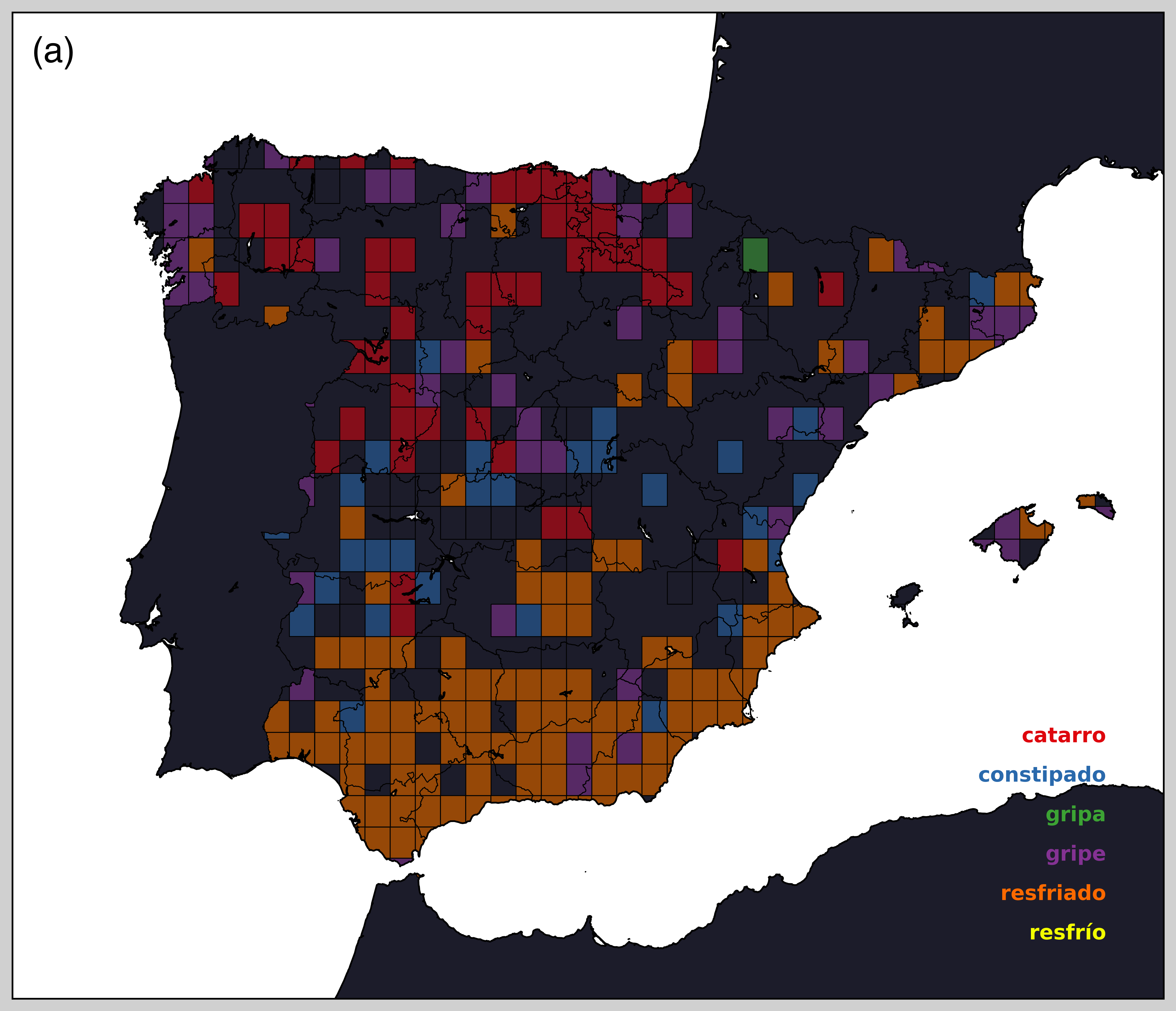}}
{\includegraphics[width=.45\textwidth]{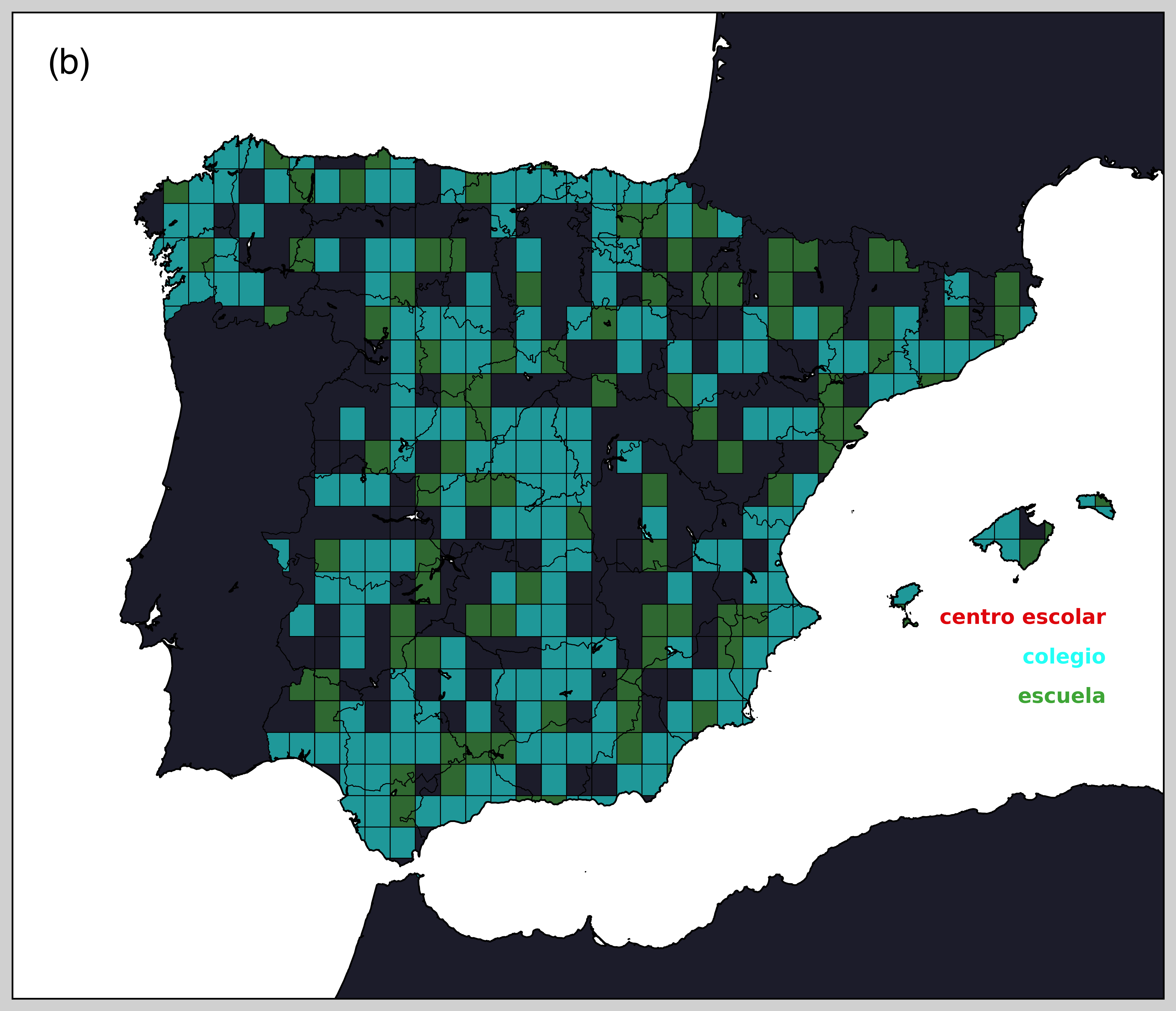}}
{\includegraphics[width=.45\textwidth]{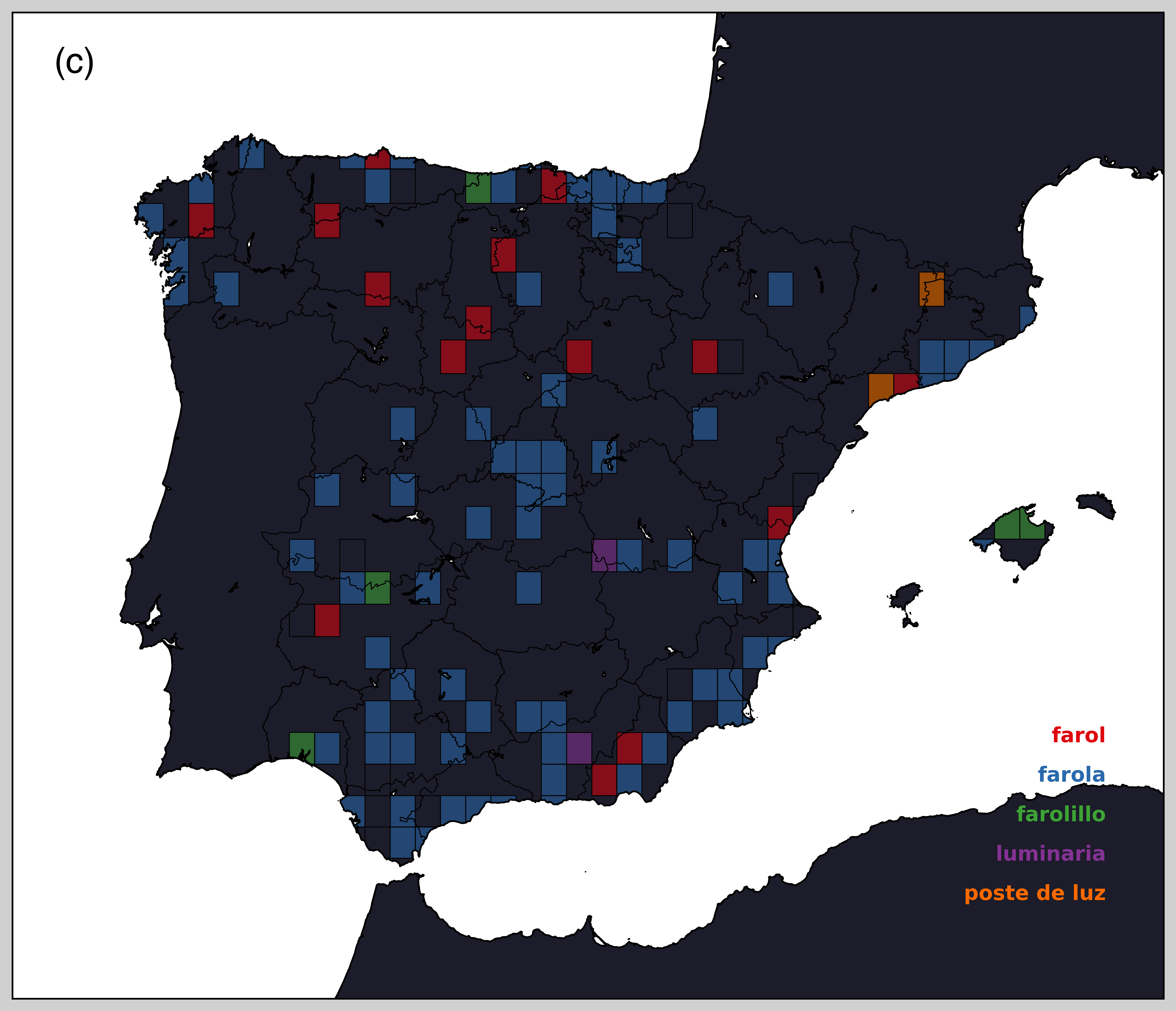}}
{\includegraphics[width=.45\textwidth]{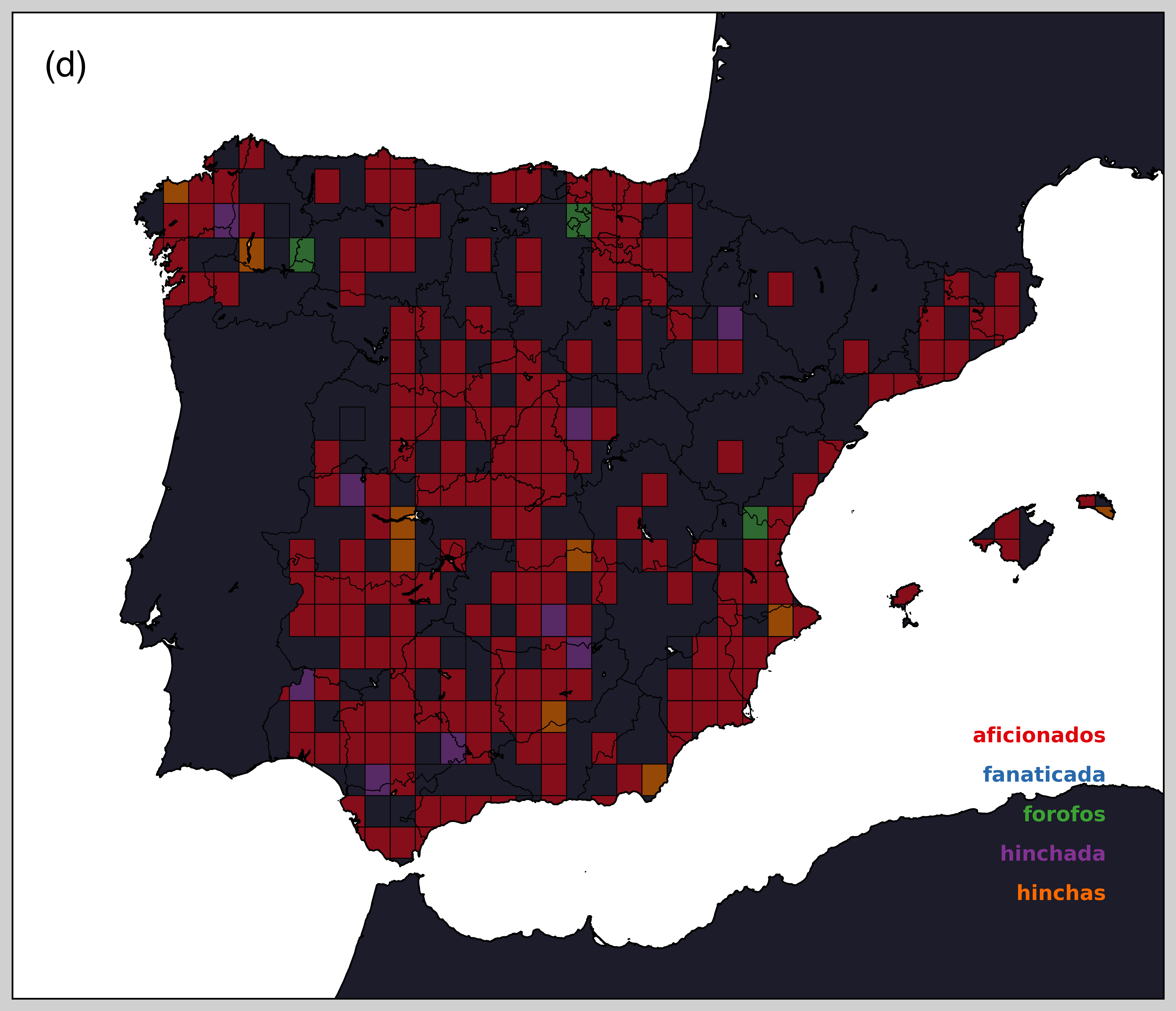}}
\caption{Spatial distribution of a few representative concepts based on the maximum absolute frequency criterion.
Each concept has a lexical variation as indicated in the figure.
The concepts are: (a) \emph{cold}, (b) \emph{school}, (c) \emph{streetlight}, (d) \emph{fans}.}
\label{qualitative}
\end{figure*}

The pictorial representation of these concepts is made using a shapefile of both the Iberian Peninsula and the Balearic Islands. Then, we construct a polygon grid over the shapefile. The size of the cells ($0.35^\circ \times 0.35^\circ$) roughly corresponds to $1200$~km$^2$. We locate the cell in which a given keyword matches and assign a different color to each keyword. We follow a majority criterion, i.e., we depict the cell with the keyword color whose absolute frequency is maximum. This procedure nicely yields a useful geographical representation of how the different variants for a concept are distributed over the space. 

\subsection{Language distance}

The dialectometric differences are quantified between regions defined with the aid of our cells.
For this purpose we take into account two metrics, which we now briefly discuss.

\subsubsection{Cosine similarity}
This metric is a vector comparison measure. It is widely used in text classification, information retrieval and data mining~\cite{murphy}.
Let $u$ and $v$ be two vectors whose components are given by the relative frequencies of the lexical variations for a concept
within a cell. Quite generally, $u$ and $v$ represent points in a high-dimensional space. The similarity measure $d(u,v)$
between these two vectors is related to their inner product conveniently normalized to the product of their lengths,
\begin{equation}\label{cosine}
d(u,v)=1-\frac{u\cdot v}{|u||v|}\,.
\end{equation}
This expression has an easy interpretation. If both vectors lie parallel,
the direction cosine is 1 and thus the distance becomes $d=0$.
Since all vector components in our approach are positive, the upper bound of $d$ is 1,
which is attained when the two vectors are maximally dissimilar. 

\subsubsection{Jensen-Shannon metric}
This distance is a similarity measure between probability density functions~\cite{lin91}.
It is a symmetrized version of a more general metric, the Kullback-Leibler divergence. Let  $P$ and $Q$ be two probability distributions. In our case, these functions are built from
the relative frequencies of each concept variation. Our frequentist approach differs from
previous dialectometric works, which prefer to measure distances using
the Dice similarity coefficient or the Jaccard index~\cite{manning}.

The Kullback-Leibler divergence is defined as
\begin{equation}
D_{KL}(P||Q)=\sum_i P(i)\log\frac{P(i)}{Q(i)}\,.
\end{equation}
We now symmetrize this expression and take the square root,
\begin{equation}\label{jsd}
JSD(P||Q)=\sqrt{\frac{D_{KL}(P||M)+D_{KL}(Q||M)]}{2}}\,,
\end{equation}
where $M=(P+Q)/2$.
The Jensen-Shannon distance $JSD(P||Q)$ is indeed a metric, i.e., it satisfies the triangle inequality. Additionally,
$JSD(P||Q)$ fulfills the metric requirements of non-negativity, $d(x,y)=0$ if and only if $x=y$ (identity of indiscernibles)
and symmetry (by construction). This distance has been employed in bioinformatics and genome comparison \cite{sim09,itz10}, social sciences \cite{ded13} and machine learning \cite{goo14}. To the best of our knowledge,
it has not been used in studies of language variation. An exception is the work of Sanders~\shortcite{sanders},
where $JSD$ is calculated for an analysis of syntactic variation of Swedish. Here, we propose to apply the Jensen-Shannon
metric to \textit{lexical} variation. Below, we demonstrate that this idea leads to quite promising results.

\subsubsection{Average distance}

Equations~\ref{cosine} and~\ref{jsd} give the distance between cells $A$ and $B$ for a certain concept.
We assign the global linguistic distance in terms of lexical variability between two cells to the mean value
\begin{equation}\label{average}
D(A,B)=\frac{\sum_i d_i(A,B)}{N}\,,
\end{equation}
where $d_i$ is the distance between cells $A$ and $B$ for the $i$-th concept and $N$ is the total number of concepts used to compute the distance. In the cosine similarity model, we replace $d_i$ in equation~\ref{average} with equation~\ref{cosine} whereas in the Jensen-Shannon metric $d_i$ is given by equation~\ref{jsd}.

\section{Results and discussion}

We first check the quality of our corpus with a few selected concepts. Examples of their spatial distributions
can be seen in figure~\ref{qualitative}. The lexical variation depends on the particular concept and on the keyword frequency.
We recall that the majority rule demands that we depict the cell with the color corresponding to the most popular word.
Despite a few cells appearing to be blank, we have instances in most of the map. 
Importantly, our results agree with the distribution for the concept \emph{cold} reported by Gon\c{c}alves and S\'anchez \shortcite{BD}
with a different corpus. The north-south bipartition of the variation suggested in figure~\ref{qualitative}(a)
also agrees with more traditional studies~\cite{fernandez}. As a consequence, these consistencies support the validity of our data.
The novelty of our approach is to further analyze this dialect distribution with a quantitative measure as discussed below.

\subsection{Single-concept case}

\begin{figure}[t]
\centering
{\includegraphics[width=.48\textwidth]{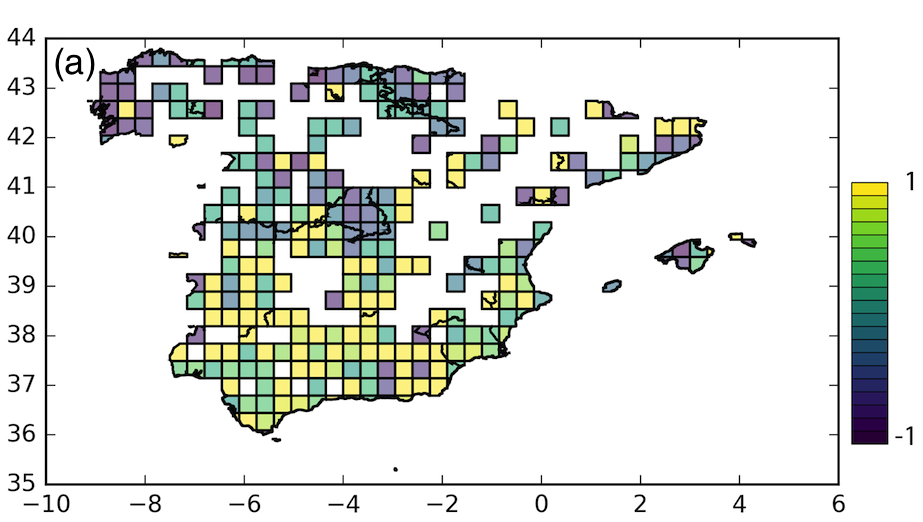}}
{\includegraphics[width=.48\textwidth]{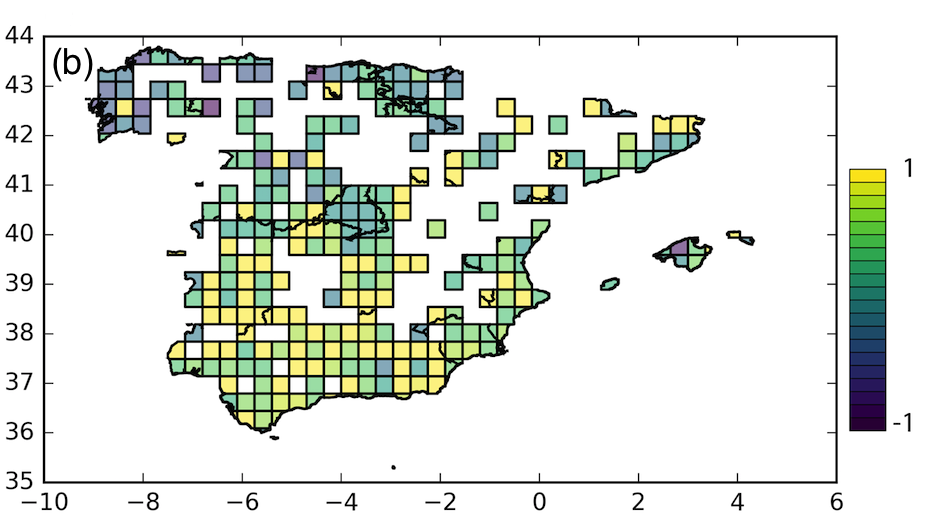}}
\caption{Linguistic distances for the concept \emph{cold} using (a) cosine similarity and (b) Jensen-Shannon divergence metrics.
The horizontal (vertical) axis is expressed in longitude (latitude) coordinates.}
\label{wothreshold}
\end{figure}

Let us quantify the lexical difference between regions using the concept \emph{cold} as an illustration.
First, we generate a symmetric matrix of linguistic distances $m_{ij}(d)$ between pairs of cells $i$ and $j$
with $d$ calculated using equation~(\ref{cosine}) or equation~(\ref{jsd}).
Then, we find the maximum possible $d$ value in the matrix ($d_{\rm max}$)
and select either its corresponding $i_{\rm max}$ or $j_{\rm max}$
index as the reference cell. Since both metrics are symmetric, the choice between $i_{\rm max}$ and $j_{\rm max}$
should not affect the results much (see below for a detailed analysis).
Next, we normalize all values to $d_{\rm max}$ and plot the distances to the reference cell
using a color scale within the range $[-1,1]$, whose lowest and highest values are set for convenience due to the normalization procedure.
The results are shown in figure~\ref{wothreshold}. Panel (a) [(b)] is obtained with the cosine similarity (Jensen-Shannon metric).
Crucially, we observe that both metrics give similar results, which confirm the robustness of our dialectometric method.

\begin{figure}[t]
\centering
{\includegraphics[width=.48\textwidth]{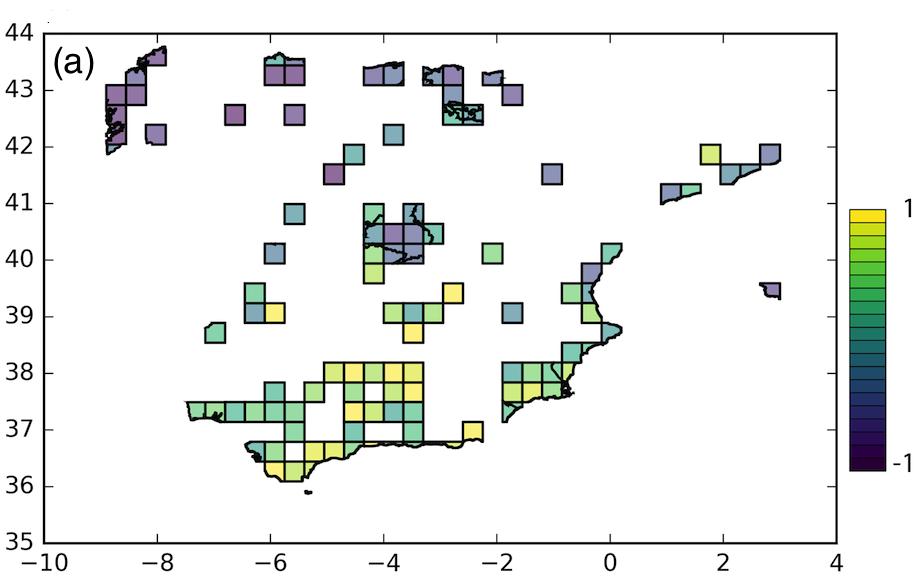}}
{\includegraphics[width=.48\textwidth]{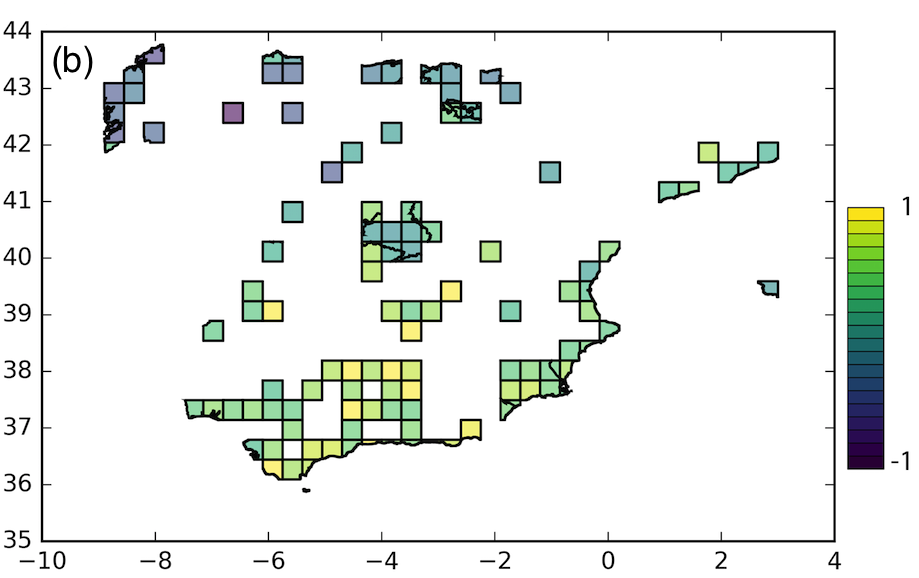}}
\caption{Linguistic distances as in figure~\ref{wothreshold} but with a minimum threshold of 5 tweets in each cell using (a) cosine similarity and (b) Jensen-Shannon metric.}
\label{threshold5}
\end{figure}

\begin{figure}[t]
\centering
{\includegraphics[width=.48\textwidth]{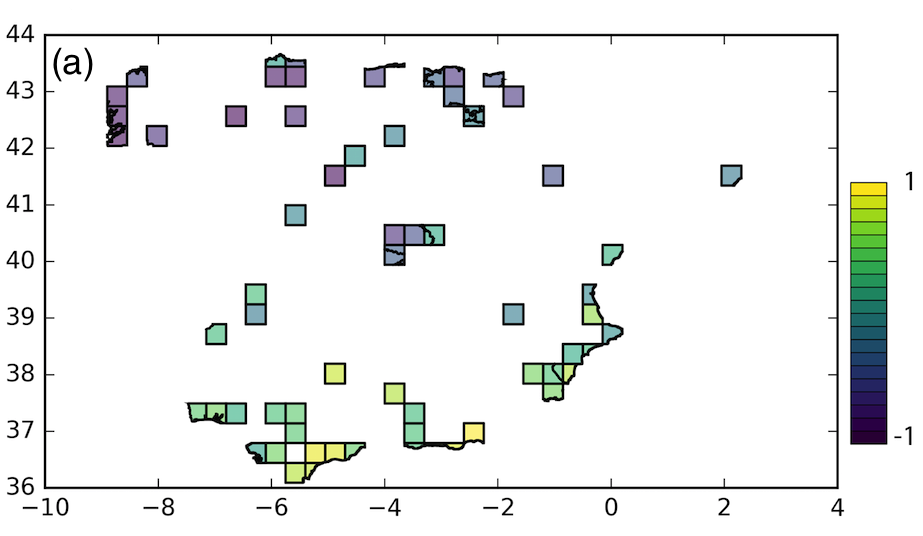}}
{\includegraphics[width=.48\textwidth]{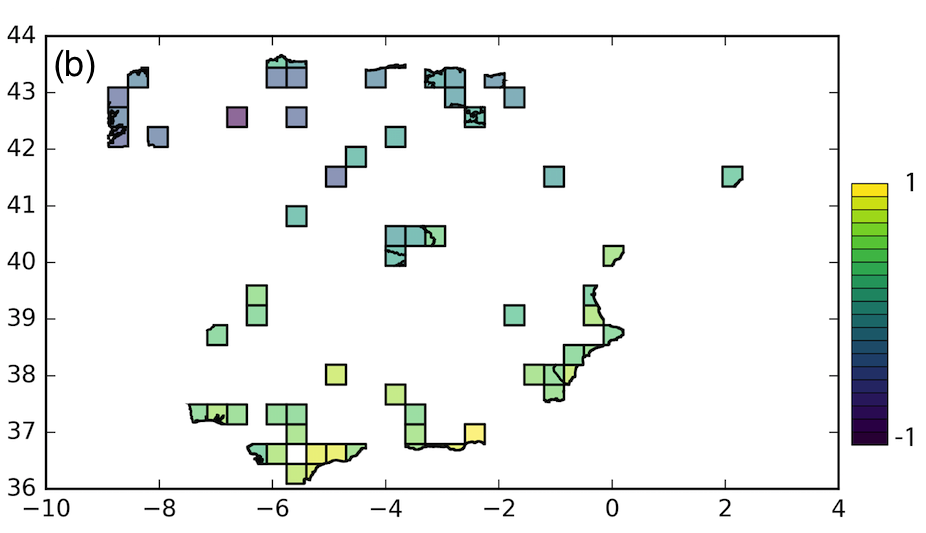}}
\caption{Linguistic distances as in figure~\ref{wothreshold} but with a minimum threshold of 10 tweets in each cell using (a) cosine similarity and (b) Jensen-Shannon metric.}
\label{threshold10}
\end{figure}

Clearly, cells with a low number of tweets will largely contribute to fluctuations in the maps. To avoid this noise-related effect,
we impose in figure~\ref{threshold5} a minimum threshold of 5 tweets in every cell. Obviously, the number of colored cells
decreases but fluctuations become quenched at the same time. If the threshold is increased up to 10 tweets,
we obtain the results plotted in figure~\ref{threshold10}, where the north-south bipartition is now better seen.
We stress that there exist minimal differences between the cosine similarity and the Jensen-Shannon metric models.

\subsection{Global distance}

\begin{figure}[t]
\centering
{\includegraphics[width=.48\textwidth]{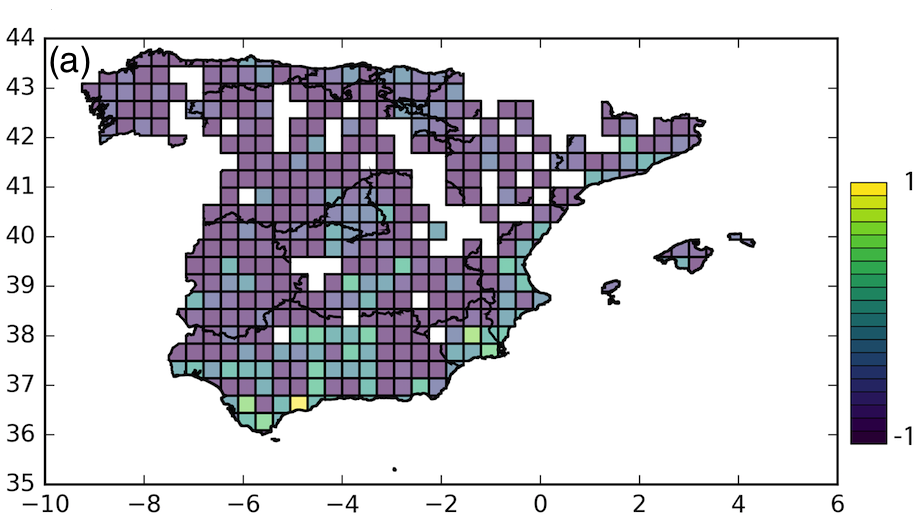}}
{\includegraphics[width=.48\textwidth]{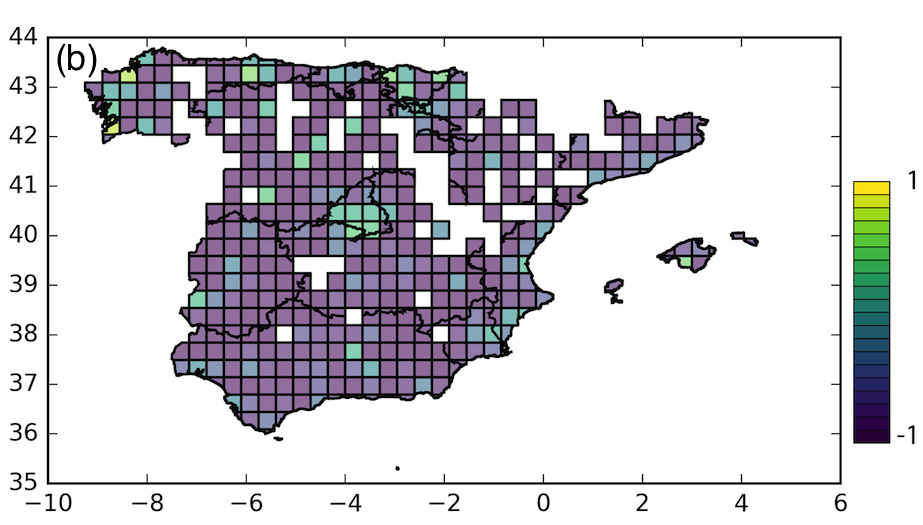}}
\caption{Global distances averaged over all concepts. Here, we use the cosine similarity measure to calculate the distance. The color distribution displays a small variation from (a) to (b) due to the change of the reference cell.}
\label{totalcosine}
\end{figure} 

\begin{figure}[t]
\centering
{\includegraphics[width=.48\textwidth]{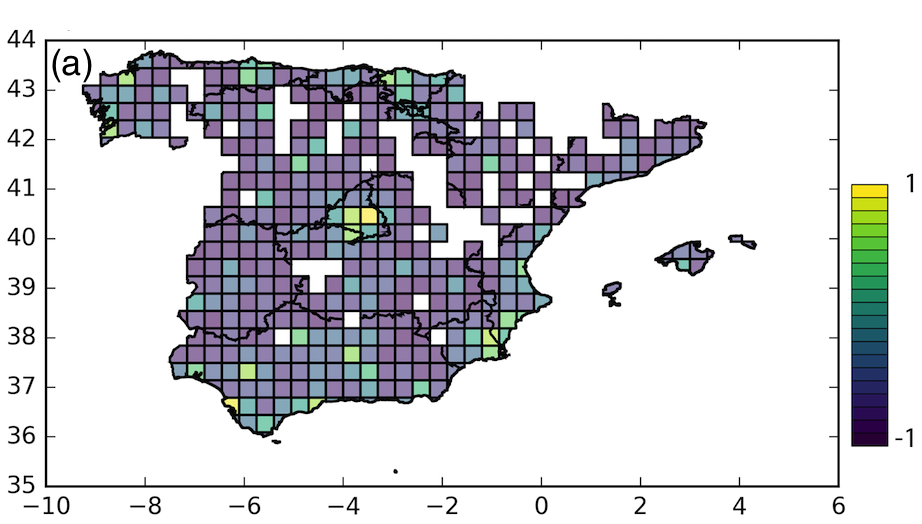}}
{\includegraphics[width=.48\textwidth]{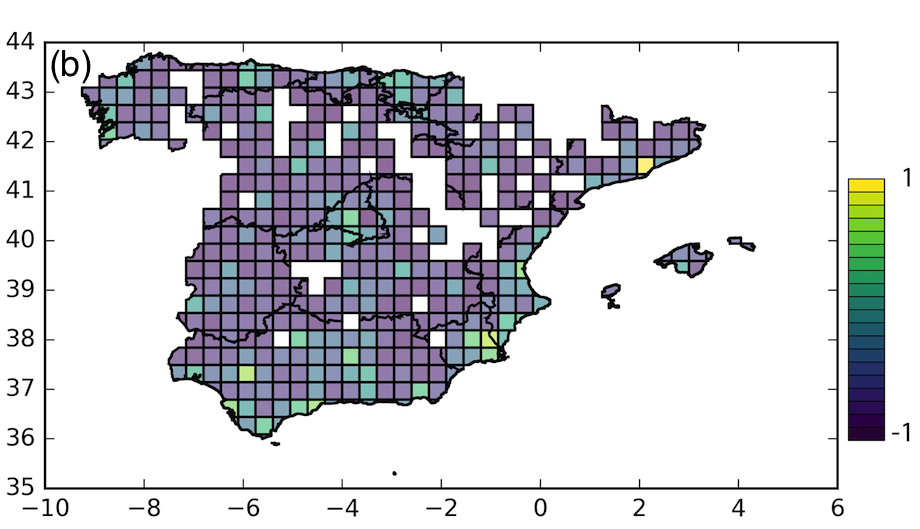}}
\caption{Global distances averaged over all concepts. Here, we use the Jensen-Shannon metric to calculate the distance. The color distribution displays a small variation from (a) to (b) due to the change of the reference cell.}
\label{totaljsd}
\end{figure}

Our previous analysis assessed the lexical distance for a single concept (\emph{cold}). Let us now take into account all concepts
and calculate the averaged distances using equation~(\ref{average}). To do so, we proceed as above and
measure the distance from any of the two cells that presents the maximal value of $d$, where $d$ is now calculated from
equation~\ref{average}. As aforementioned, $d_{\rm max}$ connects two cells, which denote as $C_1$ and $C_2$. Any of these can be selected as the reference cell from which the remaining linguistic distances are plotted in the map.
To ensure that we obtain the same results, we plot the distance distribution in both directions.
The results with the cosine similarity model are shown in figure~\ref{totalcosine}.
It is worth noting that qualitatively the overall picture is only slightly modified when the reference cell is changed
from $C_1$ [figure~\ref{totalcosine}(a)] to $C_2$ [figure~\ref{totalcosine}(b)].
The same conclusion is reached when the distance is calculated with the Jensen-Shannon metric model,
see figures~\ref{totaljsd}(a) and (b).


After averaging over all concepts, we lose information on the lexical variation that each concept presents but on the other hand one
can now investigate which regions show similar geolectal variation, yielding well defined linguistic varieties.
Those cells that have similar colors in either figure ~\ref{totalcosine} or figure~\ref{totaljsd} are expected to be ascribed to the same
dialect zone. Thus, we can distinguish two main regions or clusters in the maps. The purple background covers most of the map
and represents rural regions with small, scattered population. Our analysis shows that this group of cells possesses
more specific words in their lexicon. In contrast, the green and yellow cells form a second cluster that is largely concentrated
on the center and along the coastline, which correspond to big cities and industrialized areas. In these cells,
the use of standard Spanish language is widespread due probably to school education, media, travelers, etc.
The character of its vocabulary is more uniform as compared with the purple group. While the purple cluster
prefer particular utterances, the lexicon of the urban group includes most of the keywords.
Importantly, we emphasize that both distance measures (cosine similarity and Jensen-Shanon)
give rise to the same result, with little discrepancies on the numerical values that are not significant.
The presence of two Twitter superdialects (urban and rural) has been recently suggested~\cite{BD}
based on a machine learning approach. Here, we arrive at the same conclusion but with a totally distinct model and corpus.
The advantage of our proposal is that it may serve as a useful tool for dialectometric purposes.

\section{Conclusions}
To sum up, we have presented a dialectrometric analysis of lexical variation in social media posts
employing information-theoretic measures of language distances. We have considered a grid of cells in Spain and have calculated
the linguistic distances in terms of dialects between the different regions.
Using a Twitter corpus, we have found that the synchronic variation of Spanish can be grouped into two types of clusters. The first region shows more lexical items and is present in big cities. The second cluster corresponds to rural regions, i.e., mostly villages and less industrialized regions. Furthermore, we have checked that the different metrics used here lead to similar results in the analysis of the lexical variation for a representative concept and provide a reasonable description to language variation in Twitter.

We remark that the small amount of tweets generated after matching the lexical variations of concepts within our automatic corpus puts a limit to the quantitative analysis, making the differences between regions small. Our work might be improved by similarly examining Spanish tweets worldwide, specially in Latin America and the United States. This approach should give more information on the lexical variation on the global scale and would help linguists in their dialectal classification work of micro- and macro-varieties. Our work hence represents a first step into the ambitious task of a thorough characterization of language variation using big data resources and information-theoretic methods.

\section*{Acknowledgments}

We thank both F. Lamanna and Y. Kawasaki for useful discussions
and the anonymous reviewers for nice suggestions.
GD acknowledges support from the SURF@IFISC program.

\bibliography{vardial2017_DS}
\bibliographystyle{eacl2017}

\section*{Supplementary material}

Here we provide a list of our employed concepts and their lexical variants.

\begin{table}[H]
\begin{center}
\begin{tabularx}{\linewidth}{l X}
\hline \bf Concept & \bf Keywords \\ \hline
stapler & abrochador, abrochadora, clipiador, clipiadora, clipsadera, corchetera, cosedora, engrampador, engrampadora, engrapador, engrapadora, grapadora, ponchadora, presilladora \\
sidewalk & acera, andén, badén, calzada, contén, escarpa, vereda \\
bedspread & acolchado, colcha, colchón, cubrecama, cubrecamas, cubrelecho, edredón, sobrecama \\
flight attendant & aeromoza, azafata, hostess, stewardess \\
poster & afiche, anuncio, cartel, cartelón, letrero, póster, propaganda, rótulo, tablón de anuncio \\
pencil sharpner & afilalápices, afilalápiz, afilaminas, maquineta, sacapunta, sacapuntas, tajador, tajalápices, tajalápiz \\
bra & ajustador, ajustadores, brasiel, brassiere, corpiño, portaseno, sostén, soutien, sutién, sujetador, tallador \\
swimming pool & alberca, pileta, piscina \\
\hline
\end{tabularx}
\end{center}
\end{table}

\makeatletter
\setlength{\@fptop}{0pt}
\setlength{\@fpbot}{0pt plus 1fil}
\makeatother

\begin{table}[t]
\begin{center}
\begin{tabularx}{\linewidth}{l X}
\hline \bf Concept & \bf Keywords \\ \hline
popcorn & alepa, cabritas de maíz, canchita, canguil, cocaleca, cotufas, crispetas, crispetos, maíz pira, palomitas, pipocas, pochocle, pochoclo, pocorn, popcorn, poporopo, pororó, rosita de maíz, tostones \\
sandals & alpargata, chanclas, chancletas, chinelas, cholas, cutalas, cutaras, pantuflas, sandalias, zapatillas \\
aluminum paper & alusa-foil, foil, papel albal, albal, papel reinolds, papel aluminio, papel de aluminio, papel de estaño, papel de plata, papel encerado, papel estañado, papel para cocinar, papel platina \\
glasses & anteojos, espejuelos, gafas, gafotas, lentes \\
store window & aparador, escaparate, mostrador, vidriera, vitrina \\
coat hanger & armador, cercha, colgador, gancho de ropa, percha, perchero \\
elevator & ascensor, elevador \\
headphones & audífonos, auriculares, cascos, casquitos, headphones, hédfons, talquis \\
car & auto, automóvil, carro, coche, concho, movi \\
bus & autobús, autocar, bus, camioneta, guagua, microbús, ómnibus, taxibús \\
jeans & azulón, azulones, blue jean, bluyín, blue jeans, bluyíns, jeans, yíns, lois, mahón, mahones, pantalón de mezclilla, pantalones de mezclilla, pantalón vaquero, pantalones vaqueros, pantalones tejanos, vaqueros, tejanos, pitusa, pitusas \\
backpack & backpack, bolsón, macuto, mochila, morral, salveque \\
boat & barca, bote, canoa, cayuco, chalana, lancha, patera, yola \\
\hline
\end{tabularx}
\end{center}
\end{table}

\begin{table}[t]
\begin{center}
\begin{tabularx}{\linewidth}{l X}
\hline \bf Concept & \bf Keywords \\ \hline
fender & barrero, cubrebarro, cubrerruedas, guardabarro, guardafango, guardalodo, guardalodos, guardapolvo, lodera, polvera, quitalodo, salpicadera, salpicadero, tapabarro \\
sandwich & bocadillo, bocadito, bocata, emparedado, sandwich, sangüis, sangüich, sanwich \\
suitcase & bolso de viaje, maleta, valija, veliz \\
boxers & bombacho, bóxers, calzón, calzoncillo, calzoncillos, pantaloncillos, ropa interior, slip, trusa, taparrabos, jokey \\
lighter & bricke, brík, chispero, encendedor, fosforera, lighter, láiter, mechero, yesquero, zippo \\
backhoe & buldózer, buldócer, caterpillar, caterpílar, excavadora, máquina excavadora, maquina topadora, moto-pala, pala excavadora, pala mecánica, retroexcavadora, topadora \\
pot/pan & cacerola, cacico, cacillo, caldero, cazo, cazuela, olla, paila, pota, tartera, cazuela, sartén, freidera, freidero, fridera, paila \\
socks & calcetas, calcetines, medias, soquetes \\
matchstick & cerilla, cerillo, fósforo \\
reclining chair & cheilón, butaca, camastro, catre, cheslón, gandula, hamaca, perezosa, repo, reposera, silla de extensión, silla de playa, silla de sol, silla plegable, silla plegadiza, silla reclinable, tumbona \\
living room & comedor, cuarto de estar, estancia, líving, livin, recibidor, sala de estar, salita de estar, salón \\
computer & computador, computadora, microcomputador, microcomputadora, ordenador, PC \\
\hline
\end{tabularx}
\end{center}
\end{table}

\begin{table}[t]
\begin{center}
\begin{tabularx}{\linewidth}{l X}
\hline \bf Concept & \bf Keywords \\ \hline
headlight & cristal de frente, cristal delantero, luna delantera, lunas delanteras, luneta, parabrisa, parabrisas, vidrio delantero, windshield \\
skirt & enagua, falda, pollera, saya \\
blackboard & encerado, pizarra, pizarrón, tablero \\
dish drainer & escurreplatos, escurridero, escurridor, platera, secaplatos, secavajilla \\
poncho & estola, jorongo, mañanera, poncho, ruana \\
street light & farol, farola, farolillo, luminaria, poste de luz, poste eléctrico \\
dishwasher & friegaplatos, lavadora de platos, lavaloza, lavaplatos, lavatrastos, lavavajilla, lavavajillas, máquina de lavar platos \\
refrigerator & frigorífico, heladera, hielera, nevera, refrigerador, refrigeradora \\
toilet paper & papel confórt, papel confor, papel de baño, papel de inodoro, papel de water, papel de váter, papel higiénico, papel sanitario, papel toalet, rollo de papel \\
record player & wurlitzer, burlítser, chancha, compactera, gramola, juke box, máquina de música, pianola, rocola, tragamonedas, roconola, sinfonola, tocadiscos, traganíquel, vellonera, vitrola \\
slice of cheese  & lámina de queso, lasca de queso, loncha de queso, lonja de queso, rebanada de queso, rodaja de queso, slice de queso, tajada de queso, queso de sandwich, queso en lonchas, queso en rebanadas, queso en slice, queso americano, tranchetes  \\
demijohn & bid\'on, bombona, botella grande, garrafa, garraf\'on, tambuche, candungo, pomo pl\'astico \\
washer & lavadora, lavarropa, lavarropas, máquina de lavar \\
\hline
\end{tabularx}
\end{center}
\end{table}

\begin{table}[t]
\begin{center}
\begin{tabularx}{\linewidth}{l X}
\hline \bf Concept & \bf Keywords \\ \hline
plaster & banda adhesiva, curita, esparadrapo, tirita \\
attic & ático, altillo, azotea, buhardilla, guardilla, penthouse, mansarda, tabanco \\
wardrobe & armario, closet, placard, ropero, guardarropas \\
bracers & breteles, bruteles, suspensores, tiradores, tirantes \\
ring & anillo, argolla, aro, sortija, cintillo \\
tape recorder & cassette, casete, grabador, grabadora, magnetofón, tocacintas, magnetófono \\
blindman’s buff  & escondidas, gallina ciega, gallinita ciega, gallito ciego, pita ciega, gallo ciego \\
merry-go-round & caballitos, calesita, carrusel, tiovivo, machina \\
loudspeaker & altavoz, altoparlante, altovoz, amplificador, megáfono, parlante, magnavoz \\
flower pot & maceta, macetero, matera, matero, tiesto, macetera, plantera \\
fans & afición, aficionados, fanáticos, fanaticada, forofos, hinchada, hinchas, seguidores \\
waiter & camarero, barman, mesero, mesonero, mozo, camarero \\
school & colegio, escuela, centro escolar, scuela \\
amusement & distracciones, diversión, entretención, entretenimiento, pasatiempo \\
stay & estada, estadía, estancia \\
miss & equivocación, error, falencia, fallo \\
cheek & cachetes, carrillos, galtas, mejillas, mofletes, pómulo \\
monkey & chango, chimpancé, macaco, mono, mico, simio, chongo \\
mosquito & cínife, mosco, mosquito, zancudo \\
chance & bicoca, chance, ocasión, oportunidad \\
sponsor & auspiciador, auspiciante, espónsor, patrocinador, patrocinante, propiciador, sponsor \\
\hline
\end{tabularx}
\end{center}
\end{table}

\begin{table}[t]
\begin{tabularx}{\linewidth}{l X}
\hline \bf Concept & \bf Keywords \\ \hline
parcel & encomienda, paquete postal \\
banana & banana, banano, cambur, guineo, plátano, tombo \\
dust & nube de polvo, polvadera, polvareda, polvazal, polvero, polvoreda, polvorín, terral, terregal, tierral, tolvanera \\
bar & bar, boliche, cantina, cervecería, pulpería, taberna, tasca, expendio, piquera \\
earthquake & movimiento telúrico, movimiento sísmico, remezón, seísmo, sismo, temblor de tierra, terremoto \\
shooting & abaleo, balacera, baleada, tiroteo \\
glance & ojeada, miradita, vistazo \\greasy & engrasado, grasiento, grasoso, mantecoso, seboso \\
beautiful & bella, bonita, hermosa, linda, preciosa \\
cold & catarro, constipado, coriza, gripa, gripe, resfrío, resfriado, trancazo \\
cellophane tape & celo, celofán, cinta adhesiva, cinta scotch, cintex, scotch, teip, dúrex, diurex, cinta pegante \\
crane & grúa, guinche, tecle \\
fruit cup & ensalada de frutas, macedonia, clericó, cóctel de frutas, tuttifruti, tutifruti \\
gas station & bomba de gasolina, bomba de nafta, estación de servicio, gasolinera, bencinera, bomba de bencina, gasolinería, surtidor de gasolina \\
interview & entrevistar, reportear, interviuvar \\
obstinate & cabezón, cabezudo, cabeza dura, cabezota, obstinado, porfiado, terco, testarudo, tozudo \\
peanut & cacahuate, cacahuete, maní, cacahué, cacaomani \\
scratch & arañazo, arañón, aruñetazo, aruñón, rajuño, rayón, rasguño, rasguñón \\
sweetener & edulcorante, endulzante, endulcina, endulzador, sacarina \\
thaw & descongelar, deshielar \\
miss & echar de menos, extrañar, añorar \\
park & aparcar, estacionar, parquear \\
\hline
\end{tabularx}
\end{table}

\end{document}